\documentclass[sigconf]{acmart}
\AtBeginDocument{%
  \providecommand\BibTeX{{%
    \normalfont B\kern-0.5em{\scshape i\kern-0.25em b}\kern-0.8em\TeX}}}
\usepackage{hyperref}
\usepackage{empheq}
\usepackage{float}
\usepackage{amsmath,amsfonts,amsthm,bm}
\usepackage{multirow}
\usepackage{subcaption}
\usepackage{graphicx}
\usepackage{enumitem}
\usepackage{tabularx}

\copyrightyear{2021}
\acmYear{2021}
\setcopyright{acmlicensed}\acmConference[LAK21]{LAK21: 11th International Learning Analytics and Knowledge Conference}{April 12--16, 2021}{Irvine, CA, USA}
\acmBooktitle{LAK21: 11th International Learning Analytics and Knowledge Conference (LAK21), April 12--16, 2021, Irvine, CA, USA}
\acmPrice{15.00}
\acmDOI{10.1145/3448139.3448205}
\acmISBN{978-1-4503-8935-8/21/04}

\setlist[itemize]{noitemsep, topsep=0pt}

\begin{document}

\title{Toward Semi-Automatic Misconception Discovery Using Code Embeddings}

\author{Yang~Shi}
\email{yshi26@ncsu.edu}
\affiliation{%
  \institution{North Carolina State University}
  \city{Raleigh}
  \state{North Carolina}
  \country{USA}
  \postcode{27601}
}
\author{Krupal~Shah}
\email{khshah2@ncsu.edu}
\affiliation{%
  \institution{North Carolina State University}
  \city{Raleigh}
  \state{North Carolina}
  \country{USA}
  \postcode{27601}
}
\author{Wengran~Wang}
\email{wwang33@ncsu.edu}
\affiliation{%
  \institution{North Carolina State University}
  \city{Raleigh}
  \state{North Carolina}
  \country{USA}
  \postcode{27601}
}
\author{Samiha~Marwan}
\email{samarwan@ncsu.edu}
\affiliation{%
  \institution{North Carolina State University}
  \city{Raleigh}
  \state{North Carolina}
  \country{USA}
  \postcode{27601}
}
\author{Poorvaja~Penmetsa}
\email{ppenmet@ncsu.edu}
\affiliation{%
  \institution{North Carolina State University}
  \city{Raleigh}
  \state{North Carolina}
  \country{USA}
  \postcode{27601}
}
\author{Thomas~W.~Price}
\email{twprice@ncsu.edu}
\affiliation{%
  \institution{North Carolina State University}
  \city{Raleigh}
  \state{North Carolina}
  \country{USA}
  \postcode{27601}
}

\renewcommand{\shortauthors}{Shi et al.}


\begin{abstract}

Understanding students' misconceptions is important for effective teaching and assessment. However, discovering such misconceptions manually can be time-consuming and laborious. Automated misconception discovery can address these challenges by highlighting patterns in student data, which domain experts can then inspect to identify misconceptions. In this work, we present a novel method for the semi-automated discovery of problem-specific misconceptions from students' program code in computing courses, using a state-of-the-art code classification model. We trained the model on a block-based programming dataset and used the learned embedding to cluster incorrect student submissions. We found these clusters correspond to specific misconceptions about the problem and would not have been easily discovered with existing approaches. We also discuss potential applications of our approach and how these misconceptions inform domain-specific insights into students' learning processes.

\end{abstract}



\maketitle
\section{Introduction}

An accurate understanding of students' misconceptions on a given problem is important for effective teaching and assessment of student knowledge. Researchers can improve the performance of student models by explicitly representing misconceptions \cite{liu2016}. Students can benefit from feedback on specific misconceptions, especially if those misconceptions are detected automatically (e.g. in \cite{Michalenko2017}). Moreover, instructors can improve their lectures by explicitly addressing misconceptions, and better assess student knowledge by incorporating them into assignment rubrics. However, all of these applications first require researchers to \textit{discover} what misconceptions exist in a given domain, or on a specific problem. For example, while instructors may already have rubrics to assess performance, these rubrics' items may focus on qualities of the products created by students, rather than detecting specific misconceptions in student knowledge. Discovering such misconceptions is time-consuming and laborious, requiring experts to analyze traces of student work \cite{kurvinen2016programming, sirkia2012exploring} or interviews \cite{caceffo2016developing}. Such high cost of analysis makes it difficult to discover these \textit{problem-specific} misconceptions. 

\textit{Semi-automated} misconception discovery can address these challenges by highlighting patterns in student data, which domain experts then inspect to identify misconceptions. However, existing discovery methods have focused on simple association rules \cite{Elmadani2012, Guzman2010}, rather than state-of-the-art deep learning methods. Deep learning methods have already been used to accurately assess student performance on complex tasks (e.g. essay writing \cite{kuzi2019automatic}). In doing so, they also learn to represent students' submissions as vectors, which makes it easier to visualize and cluster submissions and find meaningful patterns that correspond to students' performance. This suggests that deep learning methods \textit{may} be able to go beyond assessment and support the discovery of misconceptions.

In this work, we present a novel method for the semi-automated discovery of problem-specific misconceptions from students' program code in computing courses, using a state-of-the-art, deep learning code classification model called code2vec \cite{alon2019code2vec}. Our approach trains code2vec to predict students' success on a given problem, and then domain experts can use the visualizations produced using code2vec, along with students' scores on a grading rubric, to quickly identify clusters of students who made similar mistakes on the problem. We applied this technique to one block-based programming problem from an introductory CS course for non-majors. We find these clusters: 1) Correspond to specific misconceptions about the problem, which could be addressed with targeted feedback, and 2) Identify groups of students' misconceptions that would not have been discovered from the rubric alone, or with standard code clustering approaches. Additionally, while code2vec has proven highly accurate at classifying code on \textit{large datasets}, in this paper we investigate its effectiveness at the novel task of \textit{assessing student code} on a much \textit{smaller} dataset. We discuss potential applications of our approach (e.g. refining rubrics, informing instruction and feedback), and how these misconceptions inform domain-specific insights into students' learning processes in a computing course.
The primary contributions of this work are: (1) a novel semi-automated method for identifying problem-specific misconceptions (2) an evaluation of this method on one assignment in a introductory computer science (CS0) course, revealing insights into students' specific learning processes and challenges, and (3) an evaluation of code2vec for automated code assessment, showing that it outperforms baseline models.

\section{Related Work} 

\textbf{Misconceptions and Misconception Discovery}:
Misconceptions in computing education refer to an incorrect understanding of a concept or a set of concepts, which lead to making mistakes in writing or reading programs \cite{swidan2018programming}. They are commonplace among novice learners but are many times invisible to experts and instructors \cite{sorva2013notional} -- a phenomenon explained in part by the “expert blind spot” \cite{nathan2003expert}. Understanding misconceptions helps instructors to be more informed, and provides raw material for student modeling \cite{liu2016blending}. Prior work has explored ways to automatically detect students’ misconceptions \cite{Michalenko2017}, however, these detectors required experts to label a training set with the presence of misconceptions to train the model, which only works if we already know what misconceptions to look for. In CS education, misconceptions have been elicited through laborious manual analysis of students' program code \cite{Paul2013} or interviews \cite{kaczmarczyk2010identifying}. In addition, manual construction of ``bug libraries'' can be a difficult and time consuming task for experts \cite{Baffles1994}. Nevertheless, such manual analysis, especially on a sequence of program snapshots, has been shown to provide more insightful information on students' misconceptions. For example, Davies et al. manually analysed students’ programming trace data to discover misconceptions, and found that trace data revealed more knowledge gaps than what has been shown in their final submissions \cite{Davies2015}. Other work focused on data-driven discovery of misconceptions. For example, Guzman et al. \cite{Guzman2010} used association rule mining to semi-automatically elicit misconceptions from multiple-choice questions in a CS practice test, and then allowed instructors to filter and verify meaningful misconceptions. However, such semi-automated detection of misconceptions also requires domain experts to analyze their rules to label them as constraints, and is only focused on closed-ended multiple choice questions (MCQs), rather than open-ended programming problems.

\textbf{Automated Programming Code Assessment}:
Our approach builds on automated assessment technologies to detect misconceptions. Automated assessments help reduce instructor burden and enhance students' learning. Much of the work on automated assessment using learning algorithms has been implemented in complex domains, such as open-ended learning environments \cite{segedy2011modeling} and essay grading \cite{kuzi2019automatic}. However, most of the work in CS education domain has been limited to using test-cases for assessment (e.g. in \cite{edwards2017codeworkout}), rather than adapting these more advanced learning approaches. These test-driven assessment frameworks run a student's code with hand-authored inputs, which requires additional instructor authoring effort and is difficult to apply to some problems (e.g. those with complex input, or graphical output). Some recent work has used machine learning algorithms to predict students' learning outcomes by extracting more information for generalized and more complex problems \cite{mao2019one, azcona2019user2code2vec}. Building on this work, we leverage the code2vec model \cite{alon2019code2vec} to extract structural information for automated program assessment, which in turn forms the basis for our misconception discovery approach.

Assessment isn't just about accuracy, it's also about giving the student actionable feedback. Rubrics help us do this. Automated approaches have also been used to automatically extract rubrics. Clear rubrics provide interpretable justification of why students succeeded or failed an assessment \cite{diana2018data}. Prior work has explored automatic discovery of rubrics. For example, Zhi et al. \cite{zhi2018reducing} and Diana et al. \cite{diana2018data} automatically generated rubrics from students' submissions by identifying common features of correct solutions. Dimopoulos et al. \cite{dimopoulos2013assessing} created a Learning Management Systems plugin that enhanced rubrics with additional info about students' interaction data (e.g. forum activity). However, none of these methods address how to improve rubrics by highlighting specific student misconceptions.

\textbf{Code Analysis and Representation}:
Recent advances in deep learning models for code classification \cite{alon2019code2vec} suggest their potential for applications in education. One problem that these models solve is that neither code nor text representations can be directly translated into vector representations to be processed by machine learning algorithms. One way that natural language research solves this challenge is by using a trainable embedding matrix to represent different text contents \cite{mikolov2013distributed}. Recent approaches represent the unstructured program code in a structured format that models can use as input. Among them, the most straightforward approach is to represent code as a sequence of one-hot encoders, which does not capture any of the structural information in code \cite{allamanis2016convolutional}. In CS education, Piech et al. \cite{piech2015learning} used a neural network to learn student code embeddings -- an abstract vector representation of code -- to identify which students would benefit from specific instructor feedback. However, this model considers only the output of code and ignores important structural information. They build their model based on the input and outputs of the code to extract the code embeddings, which does not leverage the important structural information for code itself. In this paper we focus on a particular approach, code2vec \cite{alon2019code2vec}, which proposed expressive representations extracted from abstract syntax trees (ASTs), which greatly outperform prior work on code classification tasks. Recent work suggests that the embeddings produced by this approach may be particularly useful for finding patterns in \textit{student code} \cite{azcona2019user2code2vec}.  In our work, we discover student misconceptions on a similar embedding matrix to represent code contents.



\section{Method}

\begin{figure}
\includegraphics[width=0.5\textwidth]{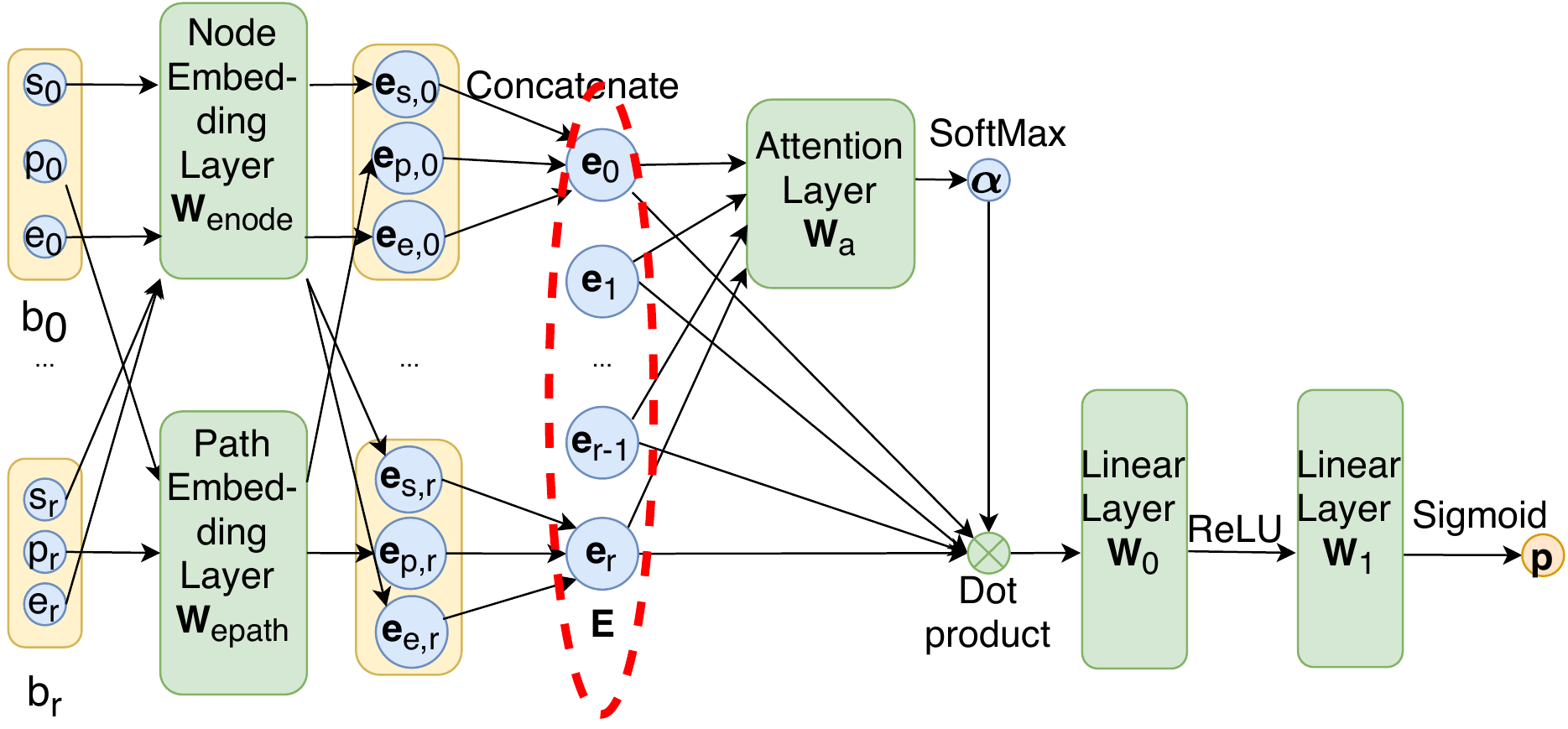}
\caption{Code2vec model. The input is represented as a set of paths $(b_0,...,b_r)$, and the model outputs the grade distribution $\bm{p}$. We extract the circled $\bm{E}$ as the embeddings for misconception discovery. The embeddings are passed through the attention layer and used for generating the prediction of students' scores on the rubric.}
\label{fig:code2vec}
\end{figure}

The code2vec model learns a code embedding during training (highlighted in Figure~\ref{fig:code2vec}), which extracts relevant structural information from students' code with respect to their success on the problem. Using this embedding, we can transform any student's code submission into a vector, which can be compared to other students' vectors. We hypothesize that this vector representation may be useful for identifying groups of students with similar misconceptions, more so than overall syntactic similarity. Embeddings are similarly used in NLP tasks \cite{mikolov2013distributed} such as sentiment classification or translation tasks. However, this vector space is sparse and large, and student submissions are embedded non-linearly in the space. We therefore leverage t-SNE \cite{maaten2008visualizing} to perform non-linear dimensionality reduction, which allows us to perform clustering and visualization more effectively. 

We perform misconception discovery only on students who unsuccessfully completed a given assignment. If a multi-item rubric is available, we can use this to make misconception discovery more effective, by discovering misconceptions among students who were unsuccessful at each individual rubric item. Since a misconception may lead students to make a specific error on an assignment, we hypothesize that students who performed poorly on a given rubric are more likely to \textit{share} a misconception. To discover these misconception groups, we use the DBSCAN algorithm  \cite{Ester96adensity-based} to cluster unsuccessful students into groups, which we hypothesize correspond to misconceptions. Not all students will have a misconception who fail a rubric item, so we use DBSCAN which can detect and remove noise point. The result is a set of clusters, which an expert can view and interpret to identify a common misconception among the submissions in each cluster.

\section{Experiment}

We set out to answer the following research questions:
\textbf{RQ1}: How accurately can a state-of-the-art code classification method assess students' performance on a programming exercise?
\textbf{RQ2}: How well does the code embedding from such a model capture meaningful similarities among students submissions, and do these reflect shared misconceptions?

\subsection{Dataset and Rubric}

Our data come from an entry-level (CS0) computer science class for non-majors at a research university, spanning four consecutive semesters (Spring 2016 through Fall 2017). Students programmed in Snap, a block-based programming environment, designed for novices \cite{garcia2015beauty}. For this experiment, we chose to analyze one assignment from the course that required the use of loops, variables and procedures to draw a spiral shape, which proved particularly difficult for students (as it has a lowest average score among all assignments). We collected $207$ submissions, which had each been graded using a rubric with 6 pass/fail items:
\begin{itemize}
    \item \textbf{Procedure with 1 Parameter}: Create a procedure to draw the shape with one parameter that controls its size.
    \item \textbf{Pen Down}: Use the ``\texttt{pen down}'' block in the procedure to start drawing.
    \item \textbf{Variable Init}: Initialize a variable and use it to control the length of each side of the shape.
    \item \textbf{Repeat Rotations}: Use a ``\texttt{repeat}'' block (i.e. a loop) to draw a spiral shape with the given size.
    \item \textbf{Forward + Turn}: Draw a square-like shape using the ``\texttt{forward}'' and ``\texttt{turn}'' blocks inside a loop.
    \item \textbf{Variable Increment}: Increment the side length variable after each iteration.
\end{itemize}

\subsection{Creating and Validating Code2Vec Embeddings}
\label{sec:baselines}

Our goal is to use code2vec to create a meaningful embedding, which can be used to detect misconceptions. For consistency, we trained the code2vec model to predict whether a student got \textit{all} rubric items correct (which 38\% of students did). For a given training set, this yielded a single embedding, which we hypothesized would reflect learned code patterns indicating success on each rubric item. However, this embedding would only be meaningful if the model itself can accurately predict student success. 

To evaluate the model's performance, we compared code2vec to three baseline models: 1) a naive majority class basline, 2) a support vector machine (SVM) (with a linear kernel, which outperformed Gaussian and Polynomial kernels), and 3) a fully connected neural network (NN). To convert student's code into a vector of numeric features to use in the baseline models, we applied term frequency–inverse document frequency (TF-IDF) to each student's code. This is similar to the Bag of Words approach used in prior learning analytics work \cite{azcona2019user2code2vec} for code feature extraction. When evaluating all models' performance, we split the Snap dataset into an $80\%$ training dataset, and a $20\%$ testing dataset. To ensure the robustness of our results in lieu of cross-validation, we re-sampled and re-trained the model $50$ times and report the average performance.

\textbf{Hyperparameters}: In training both neural network models, we set the maximum training epochs as $10,000$, with the \textit{patience} of early stopping set to $400$. We used a manual grid-search to select the learning rate ($0.1$ to $10^{-6}$), number of linear layer dimensions ($100$, $200$, $300$) and batch size ($32$, $64$, $128$, full set), and found little difference across a hyper-parameters range. We thus selected the following parameters for both neural network models. The learning rate was set to $0.0002$, linear layer dimensions were set to $100$, and the batch size was set to the full set. The weights were learned during training using the Adam optimizer \cite{kingma2014adam} for all these three models. On code2vec, we used $100$-dimensional vectors for both nodes and paths. We applied the default values from the code2vec model, since changing these numbers within a range did not meaningfully change the results. We padded the number of paths to be the same length ($100$) over the student code dataset. 


\subsection{Clustering Incorrect Submissions to Feed Misconception Detection}

We randomly selected a single code2vec model from the 50 runs and extracted its embeddings from the input of the attention layer, as shown in Fig.~\ref{fig:code2vec}. When detecting misconceptions, we applied t-SNE to visualize the \textit{entire} dataset of incorrect submissions (training and validation), since our intended use case is post hoc analytics, not predicting the misconceptions of \textit{future} students. We used DBSCAN to cluster incorrect submissions for each rubric item individually (leading to 6 \textit{sets} of clusters). For DBSCAN, we set the minimum number cluster points (minpts) at $3$, assuming that this is the smallest number of submissions an instructor or researcher might be interested in examining. We used the DMDBSCAN method \cite{elbatta2013dynamic} to determine an optimal $\epsilon$, and set $\epsilon = 11$. For each cluster, we calculated the average distance between each item in each cluster, and normalized this number by the average distance of \textit{all} points to the overall centroid. We selected up to 4 of the densest clusters for each rubric item to inspect (those with the smallest mean intra-cluster distance), which are presented in Section~\ref{sec:interpretation}.

\section{Results}
\label{sec:results}

\begin{table}
\centering
\caption{Grading accuracy, precision, recall, area under ROC curve (AUC), F-1 Score for code2vec.}
\label{table:grading_all_lak}
\begin{tabular}{|c|c|c|c|c|}
\hline
\textbf{Metric}  & \textbf{Majority} & \textbf{SVM} & \textbf{NN} & \textbf{code2vec}        \\ \hline
\textbf{Accuracy}  & 63.95\%    & 69.67\% & 71.71\%       & \textbf{75.66\% }                                   \\ \hline
\textbf{Precision}  & 0  & 61.68\% & 62.74\%         & \textbf{67.06\%}       \\ \hline
\textbf{Recall} & 0  & 42.21\% & 58.30\% & \textbf{72.66\%}               \\ \hline
\textbf{AUC} & 0.5 & 0.650 & 0.689 & \textbf{0.814}  \\ \hline
\textbf{F1} & 0  & 0.460 & 0.593 & \textbf{0.6895}  \\ \hline
\end{tabular}
\end{table}

\subsection{Model Validation: Predicting Student Performance}

Table~\ref{table:grading_all_lak} shows the mean performance of all models for predicting students' performance on the task (all correct or not). The code2vec model outperforms all baselines on all metrics. Specifically, code2vec model achieves 0.125 higher AUC, and 0.096 higher F1 score than the best among other baseline models, which is a substantial improvement. These results address RQ1, suggesting that code2vec can achieve improved assessment accuracy by extracting more meaningful features from students' code. With an accuracy of 75.7\%, this semi-automated assessment certainly cannot replace manual grading, but could reasonably \textit{assist} instructors by providing a list of likely correct submissions to be verified, or by providing formative feedback to students, warning them before submitting a potentially erroneous solution. More importantly, these results show that the structural information, encoded in code2vec embedding, is a useful predictor of student performance, more than the count information given by TF-IDF features used in the baselines. This suggests that our use of the embedding to detect meaningful patterns among student submissions is a reasonable approach.

\subsection{Embedding Interpretation}
\label{sec:interpretation}

\begin{table*}[]
\caption{Top-4 clusters Embedding Distance (ED) and Tree Edit Distance (TED), and the Interpretations of the clusters}
\label{table:codetranslation}
\begin{tabular}{|l|l|l|l|l|l|}
\hline
\textbf{Rubric}                              & \textbf{Name and Description} & \textbf{Size (\# of students)} &\textbf{Outliers} & \textbf{TED}   & \textbf{ED} \\ \hline
\multirow{4}{*}{R0 - Procedure with 1 Param.} & Fixed Shape                  & 7                 & 0 & 0.2912              & 0.1884            \\ \cline{2-6} 
                                             & Large Move/Turn                     & 11                & 1        & 1.9551      & 0.1657             \\ \cline{2-6} 
                                             & Not testing a custom block    & 3                 & 0      & 0.2868        & 0.0813             \\ \cline{2-6} 
                                             & Rotations not from Input      & 3                 & 0     & 0.4666         & 0.0788             \\ \hline
\multirow{3}{*}{R1 - Initialize Variable}    & Large Move/Turn                     & 10                & 0    & 1.9599          & 0.1680             \\ \cline{2-6} 
                                             & Medium Move/Turn                     & 4                 & 0        & 0.5730      & 0.1254             \\ \cline{2-6} 
                                             & Fail to initialize            & 3                 & 1        & 0.6041      & 0.0890             \\ \hline
\multirow{2}{*}{R2 - Move and Turn}          & Large Move/Turn                     & 6                 & 1      & 2.0339         & 0.1377            \\ \cline{2-6}
                                             & Medium Move/Turn                     & 4                 & 1      & 0.4207        & 0.1515 \\ \hline
\multirow{4}{*}{R3 - Repeat}                 & Repeat sides not rotations    & 6                  & 0   & 0.4566          & 0.1357             \\ \cline{2-6} 
                                             & Mixed incorrect repeat        & 11                  & 0      & 0.4719      & 0.2494             \\ \cline{2-6}
                                             & Mixed incorrect repeat        & 4                 & 0      & 0.9894      & 0.1342            \\ \cline{2-6}
                                             & Fixed shape                & 4                   & 1      & 0.0891      & 0.0428             \\ \hline
\multirow{2}{*}{R4 - Pen down}               & Mixed incorrect custom block  & 5                    & 0  & 0.5115        & 0.1436             \\ \cline{2-6} 
                                             & Mixed incorrect custom block  & 4                         & 0 & 0.5449      & 0.1265             \\ \hline
\multirow{2}{*}{R5 - Increment}              & Large Move/turn                     & 9                   & 0     & 1.9781       & 0.1680             \\ \cline{2-6} 
                                             & Medium Move/turn                     & 5                         & 1 & 0.5947      & 0.1481             \\ \hline
\end{tabular}
\end{table*}

\begin{figure*}
\centering

\begin{minipage}[b]{0.25\textwidth}
\includegraphics[width=\textwidth]{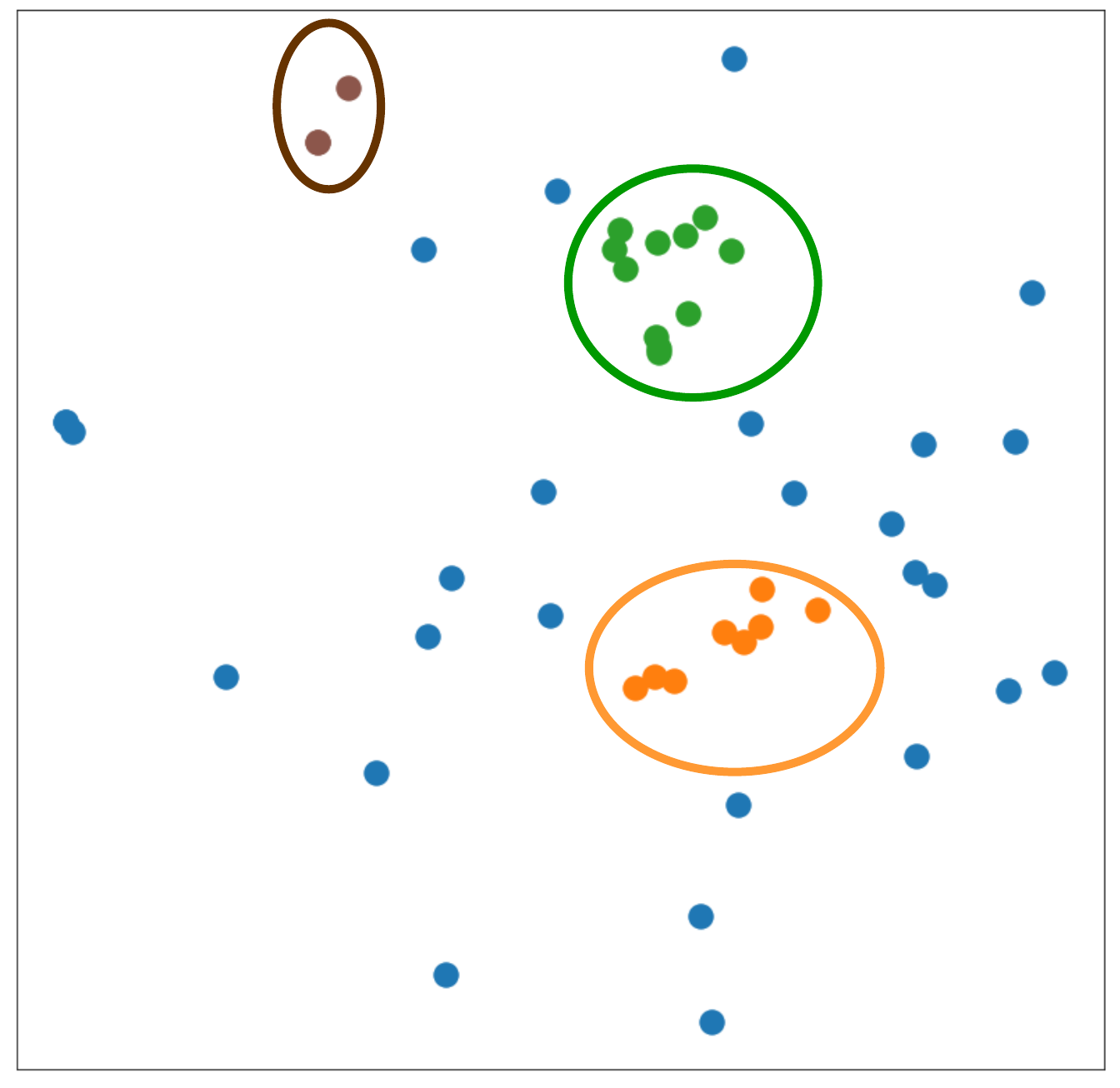}
\caption{\label{fig:embedding}Code Embedding visualized clusters from the \textbf{Procedure with 1 Input} rubric item. We colored three clusters \textbf{Rotations not from input} (Brown), \textbf{Fixed Shape} (Orange) and \textbf{Move/Turn} (Green) for discussion in Subsection~\ref{sec:casestudy}.}
\end{minipage}
\hfill
\begin{minipage}[b]{0.70\textwidth}
\includegraphics[width=\textwidth]{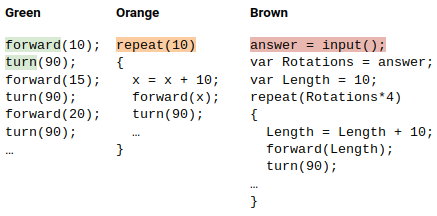}
\caption{\label{fig:pseudocode}Pseudocode examples from student code dataset for the three clustered circled in Fig.\ref{fig:embedding}. Code related with misconceptions are highlighted with specific colors.}
\end{minipage}
\end{figure*}

We introduce the interpretations of embeddings to address RQ2 in this subsection, and include a case study in the next subsection. Table~\ref{table:codetranslation} shows the 4 densest clusters discovered for each rubric item (or as many as were discovered). Recall that the primary goal of our proposed method is to discover meaningful \textit{clusters} (groups of students) who are demonstrating a similar misconception and would benefit from similar feedback. To verify this, one author, who is an instructor for the course, inspected each cluster and evaluated whether a shared misconception was present in the cluster, in which all (or nearly all) students failed the rubric item due to the same misconception. This process led to the discovery of 7 distinct misconceptions (shown in Table~\ref{table:codetranslation}), which involved viewing 63 distinct submissions (about half of the incorrect submission). The author then created a short definition for each misconception.

Of the 17 clusters identified, only 4 had no consistent pattern or misconception (no label). This suggests that our approach frequently identified meaningful groups of students. To validate the original misconception labels, 2 other authors familiar with the class inspected the clusters and marked each submission as matching or not matching the misconception, resolving any disagreements. We report then number of ``outliers'' in each cluster which did not have the misconception. Many clusters had 0 outliers, and none had more than 1, further suggesting that clusters have internal consistency. One misconception (Move/Turn) showed up across almost every rubric item, often represented by similar or identical clusters of students. In the future, this redundancy could be avoided by excluding clusters that have high overlap with previously discovered clusters. To illustrate the types of misconceptions discovered by our approach, we describe 3 of the misconceptions discovered for one rubric item in detail below.

\subsection{Case Study: ``Procedure with One Parameter" Rubric Item}
\label{sec:casestudy}

The first rubric item (referred to as \textit{R0}) requires students to create a procedure with one parameter that holds the number of rotations of the spiral-square shape. This rubric was originally designed to test students’ ability to create procedures with parameters, which demonstrates an understanding of abstraction. While the rubric item was designed to assess a single student skill, we found that students failed the item for a wide variety of reasons, as illustrated by 3 distinct misconceptions:

The \textbf{green} cluster in Figure~\ref{fig:embedding} contains 11 students who shared a misconception that iteration should be accomplished by duplicating code multiple times, (e.g. 5 pairs of “\texttt{move}” and “\texttt{turn}” blocks), rather than correctly repeating iterations with a loop. This misconception has little to do with \textit{R0} or with procedures (some students used a procedure; others did not), but it caused students to fail \textit{R0}, since they could not correctly use the parameter without a loop.
Unlike the other \textit{R0} clusters, these students needed basic feedback on a prerequisite programming concept (iteration), and would likely not benefit from feedback on procedures or parameters.
These students all failed other rubric items as well, and this misconception shows up consistently across rubric items. 

The \textbf{orange} cluster in Figure~\ref{fig:embedding} consists of 8 students who created a loop that repeats a \textit{fixed} number of times, rather than using the procedure's parameter to determines the number of rotations. This suggests a misconception about using loops with parameters. Unlike the previous cluster, students in this cluster knew how to create a loop, but they likely had a misconception that loops always iterated a \textit{fixed} number of times, which was reflected in their inability to use a parameter or even a variable in the ``\texttt{repeat}'' block. Despite this similarity, there was also variety within the cluster: two of these students used a procedure without a parameter, while six did not create a procedure, and some used variables while others did not. Despite these syntactic differences, these submissions were positioned very close together in the embedding.

The \textbf{brown} cluster in Figure~\ref{fig:embedding} consists of three students, where all of them used \textit{a local variable} to determine the number of \textit{rotations}, instead of the \textit{procedure's parameter}. In particular, two students created a procedure without parameters, then used an \texttt{ask} block to ask the user for an input and finally store it in a different created variable: ``Rotations''. The third student did not create a procedure, and created a variable: ``Rotations'' and initialized it with a static value. Unlike the previous clusters, all students demonstrated an understanding that the size of the shape should be dynamic, using a variable in the \texttt{repeat} block. However, they shared a misconception about how a \textit{procedure parameter} can be used to vary this value. This could also be explained by a misunderstanding about the assignment requiring such a parameter, but in either case, students would benefit from targeted feedback on this misconception.

\subsection{Are Clusters Distinct and Non-obvious?}
Our proposed method is only useful if it detects clusters of student code that could not easily be otherwise identified. For example, if the clusters simply correspond to getting specific items in the rubric incorrect, there would be no need for a new approach to discover them. However, we found that the individual rubric scores varied widely within each cluster, without consistent patterns of which rubric items were gotten correct or incorrect (beyond the initial rubric item used to discover the cluster, which all students got incorrect). 

Another possibility is that the clusters simply have similar ASTs, allowing existing code clustering methods to detect them. However, investigation of the code snapshots within each cluster revealed wide variety in the size and syntactic structure. To verify this, we calculated the average tree edit distance between each item in each cluster, and normalized this number by the average distance to the overall centroid. We compared this normalized cluster tree edit distance (TED) with the normalized embedding distance (ED), as shown in  Table~\ref{table:codetranslation}. We found the embedding distance is always much smaller than the tree edit distance, showing that the clustered submissions were relatively much closer together in the embedded space (compared to non-clustered submission), than in TED space, suggesting they would not be easily detected by traditional clustering methods. This shows the the embedding captures some aspects of code similarity which not apparent by simply comparing ASTs.

\section{Discussion}
\label{sec:discussion}

\textbf{Implications for Research and Teaching}:
We envision a number of potential applications of our approach to help researchers and instructors to address students' misconceptions.
\textit{Feedback propagation:} We found that students in the clusters we discovered would have benefited from targeted feedback that addressed their misconception. One application of our approach is in feedback propagation \cite{Head2017}, where an instructor can write a single piece of feedback and send it to all students who might benefit. Moreover, our clusters would not have been discovered through traditional code clustering methods used in prior work (e.g. canonicalization \cite{glassman2015overcode}).
\textit{Semi-automated feedback:} Many of the misconceptions we discovered would be best addressed with \textit{formative} feedback, as the student works. Since the patterns we discovered were relatively straightforward, one could imagine building a system to detect them automatically and offer appropriate immediate feedback early in a students' work. In our investigation, we found that our clusters were not exhaustive, and did not include all students who shared the common misconception, just a representative subset. However, prior work suggests that, once a misconception is identified, models can be trained to accurately identify it automatically \cite{Michalenko2017}.
\textit{Rubric refinement:} We found that a single rubric item can easily correspond to numerous, distinct challenges faced by students, as described in Section~ \ref{sec:casestudy}. Instructors and researchers may want to assess student knowledge in a more granular way, and our approach could help develop rubrics that capture the full range of students' ability and knowledge. For example, it might be best to split the rubric item \textit{R0} into multiple items, or convert it from a binary pass/fail item to having scale values representing the various degrees of success.

\textbf{Domain-Specific Implications for Computing Education}: 
Another application of misconception discovery is to inform our understanding of a learning domain, and here we reflect on the implications of our findings for computing education. In this paper, we focused on \textit{R0}, which assessed students' ability to use a procedure parameter to vary the size of the shape they were drawing (determined by the number of iterations in a loop). While this rubric item was designed to assess a single concept (use of a parameter), our method discovered three clusters representing \textit{distinct barriers} that students faced, requiring different remediation. The first cluster included students who had not figured out how to use a loop to draw the shape, suggesting the need for basic help on programming concepts needed in the assignment. The second cluster included students who \textit{were able} to use a loop to draw the shape, but they did not understand that the \textit{size} of the shape should \textit{vary}; they used a static value instead. These students may need help understanding the assignment requirements, or which property of the shape should be parameterized. The third cluster of students \textit{did} understand that the size of the shape should vary, but used a variable instead of the procedure parameter for this, suggesting their need to understand the use of \textit{parameters}.

These results suggest implications for how to teach the lesson preceding this assignment, and how to teach procedure parameters in general. For our classroom, the instructor may want to do a more thorough review of loops and variables before the assignment (e.g. to help students in the first/green cluster). In general, our results suggest that students may struggle to use abstraction in the absence of a clear set of steps for creating a procedure with parameters. The instructor may want to teach students steps that address each of the barriers we identified, e.g.: 1) create a non-parameterized version of the procedure, which is the same every time, and make sure it works; 2) identify what aspect(s) of the procedure should be able to vary and create corresponding parameter(s); 3) replace the corresponding literal values in your code with the appropriate parameters. This aligns with a similar series of steps presented by the popular textbook on ``\textit{How to Design Programs}'' to teach abstraction \cite{felleisen2018design}.

\textbf{Limitations:}
This study is exploratory and has important limitations. First, we focused on one assignment from one course. This allowed us to discuss the misconceptions we discovered in depth, but as a result, it is not clear how well our results will generalize to other assignments. Second, our approach -- and our evaluation -- relied on the interpretations of experts to describe and label misconceptions, making the process semi-automatic. However, we argue this interpretation is a necessary part of making use of any misconceptions to verify their meaningfulness, and in our evaluation we used multiple raters to verify cluster meaning. Third, our approach yield some clusters that have no clear interpretation, and some duplicates, suggesting the need for future work to refine the process. Lastly, the discovered clusters are not exhaustive, meaning not all students who make a given misconception show up in the cluster. However, as discussed in Section ~\ref{sec:discussion}, the clusters can still be useful to researchers and instructors, prompting changes to course structure that can benefit students \textit{not} captured by the clusters.

\section{Conclusion}
\label{sec:future-work}

In this paper, we presented a method for discovering student misconceptions semi-automatically using code assessment. We found that the code2vec model achieves a better semi-automated assessment performance than baseline models, showing that useful code structural information is learned in the model for the embeddings. We then examined if the embeddings could be used to discover misconceptions, and found that the embeddings generated clusters corresponding to meaningful and distinct misconceptions. From a case study on three clusters, we found that they are novel, consistent, and cannot be discovered by traditional code clustering methods, e.g. using tree edit distance. These results suggest that deep learning approaches have exciting potential to not only assess code, but help in the identification of learners' misconceptions.

\begin{acks}
This research was supported in part by the Amazon Web Services (AWS) Cloud Credits for Research program.
\end{acks}

\bibliographystyle{ACM-Reference-Format}
\bibliography{main}
\end{document}